\pdfoutput=1
\documentclass[letterpaper, 10 pt, conference]{IEEEtran}  
\IEEEoverridecommandlockouts
\usepackage{cite}
\usepackage{amsmath}
\usepackage{amssymb}
\usepackage{amsfonts}
\usepackage{empheq}
\usepackage{graphicx}
\usepackage{epstopdf}
\usepackage{epsfig}
\usepackage{subfigure}
\usepackage{textcomp}
\usepackage{xcolor}
\usepackage{import}
\usepackage{pdftexcmds}
\usepackage{pdfcolmk}
\usepackage{hyperref}
\usepackage{algorithm,algpseudocode}
\usepackage{setspace}
\usepackage{inputenc}

\def\BibTeX{{\rm B\kern-.05em{\sc i\kern-.025em b}\kern-.08em
    T\kern-.1667em\lower.7ex\hbox{E}\kern-.125emX}}
\begin{document}

\linespread{0.9848}

\newcommand{\q}{\textit{\textbf{q}}}
\newcommand{\qdot}{\dot{\textit{\textbf{q}}}}
\newcommand{\x}{\textit{\textbf{x}}}
\newcommand{\xdot}{\dot{\textit{\textbf{x}}}}
\newcommand{\bx}{\boldsymbol{\xi}}

\newcommand{\dal}{\dot{\alpha}}
\newcommand{\fb}{\bar{f}}

\newcommand{\St}{\textbf{S}}
\newcommand{\T}{\textbf{T}}
\newcommand{\W}{\textbf{W}}
\newcommand{\J}{\textbf{J}}
\newcommand{\A}{\textbf{\text{A}}}
\newcommand{\C}{\dot{\textbf{\textit{x}}}_\text{s}}
\newcommand{\Pt}{\textbf{\text{P}}}

\newcommand{\mk}{{\color{red}$\bigstar$}}

\newtheorem{Definition}{\vspace{0em}\textbf{{Definition}}}
\newtheorem{Property}{\vspace{0em}\textbf{Property}} 


\title{\LARGE \bf
Improving Redundancy Availability: Dynamic Subtasks Modulation for Robots with Redundancy Insufficiency
\thanks{Identify applicable funding agency here. If none, delete this.}
}
\author{
Lu Chen$^{*1}$, Lipeng Chen$^{*2}$, Xiangchi Chen$^{1}$, Yi Ren$^{2}$, Longfei Zhao$^{2}$, Yue Wang$^{1}$, Rong Xiong$^{1}$
\thanks{$^{*}$First two authors contributed equally to this work.}
\thanks{$^{1}$Lu Chen, Xiangchi Chen, Yue Wang and Rong Xiong are with Zhejiang University, Zhejiang, China. {\tt\small \{lu-chen, chenxiangchi, ywang24, rxiong\}@zju.edu.cn} }%
\thanks{$^{2}$Lipeng Chen, Yi  Ren and Longfei Zhao are with Tencent, China. {\tt\small \{lipengchen,evanyren,longfeizhao\}@tencent.com} }%
}


\maketitle
\title{\LARGE \bf
Improving Redundancy Availability: Dynamic Subtasks Modulation for Robots with Redundancy Insufficiency  
}




\begin{abstract}


This work presents an approach for robots to suitably carry out complex applications characterized by the presence of multiple additional  constraints or subtasks (e.g. obstacle and self-collision avoidance) but subject to redundancy insufficiency. 
The proposed approach, based on a novel subtask merging strategy, enforces all subtasks in due course by dynamically modulating a virtual secondary task, where the task status and soft priority are incorporated to improve the overall efficiency of redundancy resolution. The proposed approach greatly improves the redundancy availability by unitizing and deploying subtasks in a fine-grained and compact manner.
We build up our control framework on the null space projection, which guarantees the execution of subtasks does not interfere with the primary task.
Experimental results on two case studies are presented to show the performance of our approach.

\end{abstract}

\section{Introduction}
\label{sec:introduction} 

\begin{figure}[!t]
        \begin{center}	
         
                \hspace{-3.5mm}			
                \subfigure[Human avoidance subtask]{
                        \label{fig1:subfig:human}  
                        \includegraphics[height=0.36 \columnwidth, angle=0]{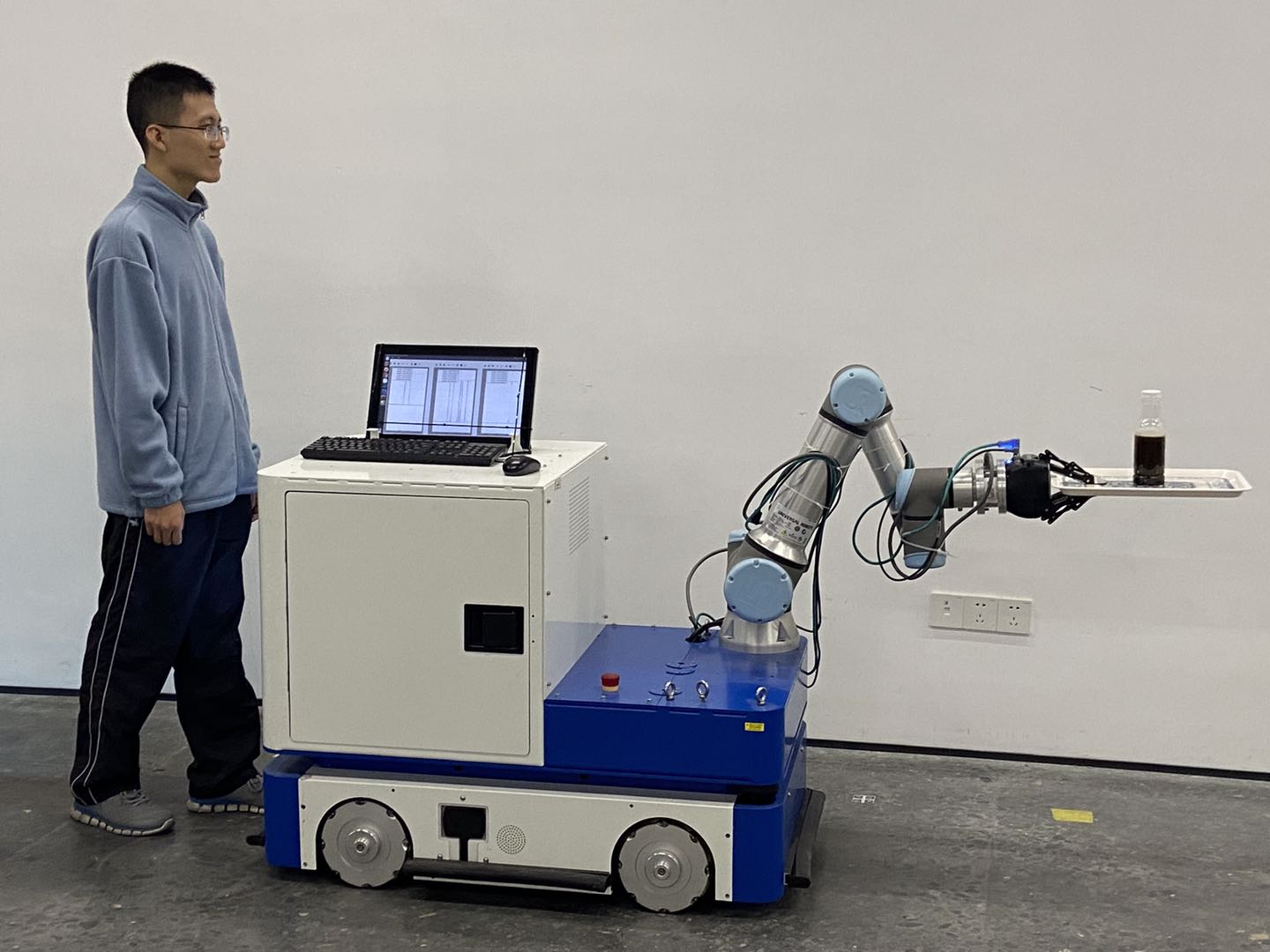}}
                \subfigure[Self-collision avoidance subtask]{{
                        \label{fig1:subfig:self}  
                        \includegraphics[height=0.36 \columnwidth, angle=0]{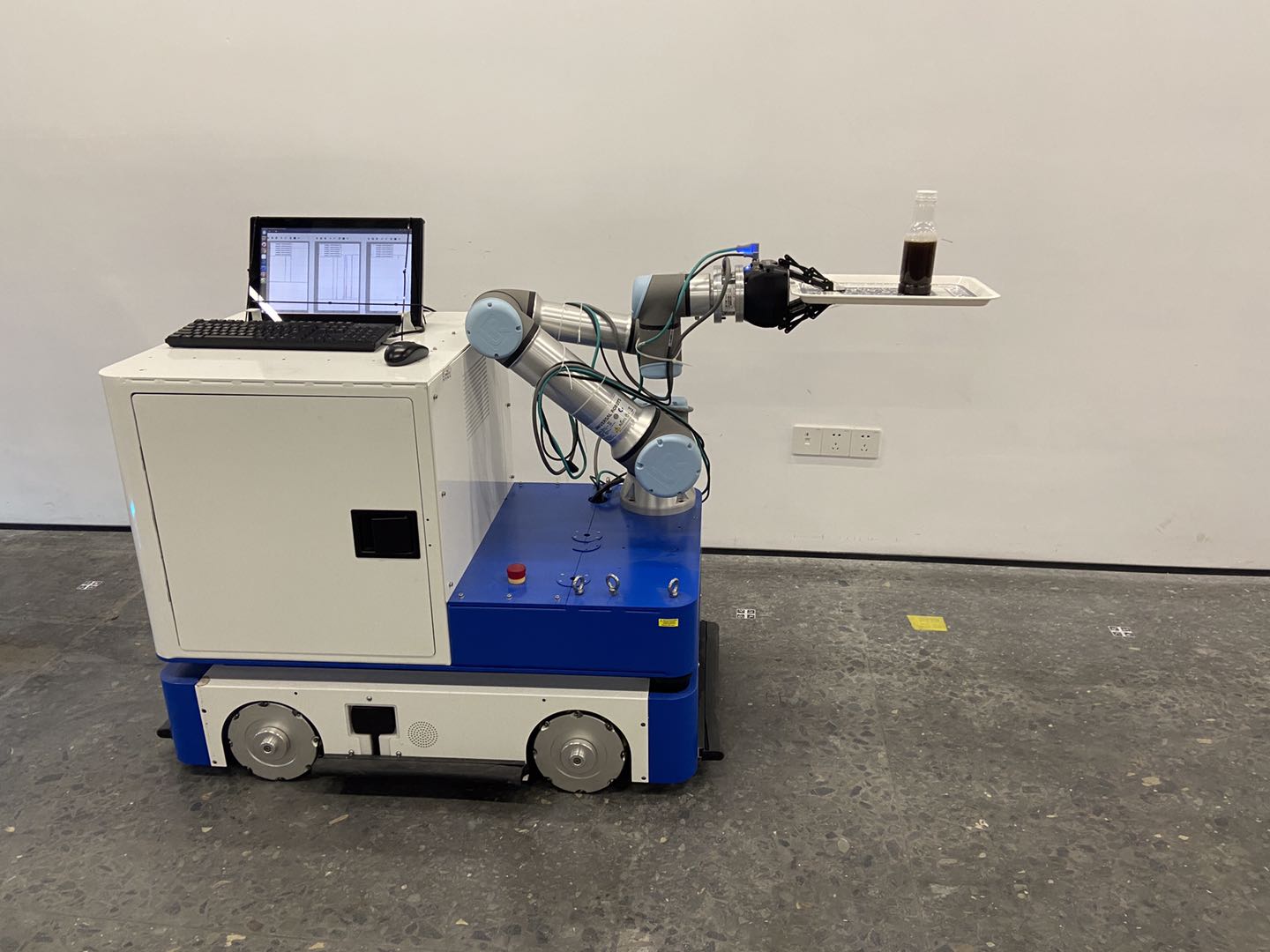}}}	
            \caption{A drink-serving robot dynamically allocates relatively insufficient redundancies to accomplish multiple subtasks while serving drinks. (a) The human avoidance subtask takes redundancies as the human walks close. (b) The self-collision subtask takes redundancies once the human is far away.}
            \label{fig1:intro}
        \end{center}
    \end{figure}

Redundant robots have been dominating with growing popularity 
the robotics community in virtue of their increased dexterity, versatility and adaptability~\cite{hanafusa1981analysis,siciliano1990kinematic,chiaverini2016redundant}.
However, except for few highly specialized systems, most redundant robots still underperform due to lack of relatively sufficient redundancies, especially when operating in unstructured or dynamic environments like households or warehouses characterized by the occurrence of multiple additional subtasks.
Take a drink-serving task as illustrated in Fig.~\ref{fig1:intro} for example. Even though the mobile robot is already equipped with nine degrees of freedom (DOF), as the robot carries a tray upright to serve drinks, only three DOFs will be left as redundancies.
However, besides the primary serving task, the robot is frequently confronted with a large number of additional constraints or subtasks, e.g. obstacles, walking humans and singularity avoidance, which may actually require far more redundancies than the remaining ones. That is, the robot may not be able to deal with all subtasks simultaneously due to the lack of redundancies for subtasking.

We focus on the constrained scenario of \textit{redundancy resolution} problems~\cite{maciejewski1985obstacle,nakamura1987task,ficuciello2015variable} like this, where a redundant robot is supposed to carry out a primary task accompanied by multiple additional subtasks \textbf{but} subject to redundancy insufficiency. 

A straightforward engineering way out of the above redundancy dilemma is to introduce more kinematic redundancies into the robot mechanical structure, which apparently is way too expensive to be repeatable. The majority of prior works on redundancy resolution, either via optimization~\cite{khatib1986real,ge2002dynamic,flacco2015discrete} or task augmentation~\cite{sciavicco1988solution,zanchettin2011general,benzaoui2010redundant}, however, are
fundamentally under the premise the robot can provide sufficient redundancies
i.e. all subtasks can be performed simultaneously with required redundancies.

Rather, we noticed that in fact not all aforementioned subtasks have to be performed simultaneously or synchronously 
thanks to task feature and environment characteristics\footnote{It is acknowledged that there exist cases where subtasks need to be performed strictly simultaneously. For such cases, the engineering augmentation as aforementioned is the only solution.}. For example, a whole-course obstacle avoidance subtask can actually be idle during most of the runtime until some obstacle appears within a certain threshold region, and therefore can be deferred from taking redundancy. Such characteristics give rise to the potential of asynchronicity among subtasks, which essentially accommodates most practical robot applications characterized by dynamic and unstructured environments. 

It leads to a lightweight but effective solution
that the robot can dynamically allocate redundancies to subtasks
according to some common rules like task urgency, activeness and importance. 
For example in Fig.~\ref{fig1:intro}, as the robot carries out the primary drinking-serving task, if a human moves closer to the robot (Fig.~\ref{fig1:subfig:human}), the subtask of human avoidance is of an increasing and ultimately dominating priority of taking all redundancies, while all other substasks will be temporarily frozen since no more redundancy is available. As the human walks away, the robot will eventually release the (part of) redundancies,
 until some other subtask takes them, e.g. the self-collision avoidance subtask (Fig.~\ref{fig1:subfig:self}). 

In this work, we borrow ideas from asynchronous time-division multiplexing (ATDM), propose an approach to subtask management for redundant robots subject to redundancy insufficiency. Our approach unfolds as follows: we first unitize all multi-dimensional subtasks to be executed along with the primary task  
into a set of one-dimensional \textit{elementary subtasks}. This step allows us to greatly improve the redundancy availability by deploying subtasks in a more fine-grained and compact manner. We then manage elementary subtasks by fusing them into a virtual multi-dimensional \textit{secondary task} w.r.t. the primary task. 
We propose a novel \text{subtask merging operator} and an efficient updating strategy to dynamically modulate the secondary task in compliance with the \textit{task status} and \textit{soft priority} derived heuristically.
Based on the approach, all subtasks can be suitably performed in due course.

Our control framework is built upon previous work of task priority based redundancy resolution~\cite{hanafusa1981analysis,maciejewski1985obstacle,nakamura1987task}, which guarantees the low-level tasks executed in the null space do not interfere with the high-level tasks. We integrate our subtask merging strategy into the null space projection technique to derive a general control framework of subtask management for redundant robots subject to redundancy insufficiency. 
In this framework, the primary task is perfectly performed using a number of required DOFs, while all other subtasks are suitably carried out as a virtual dynamic secondary task using the remaining insufficient redundancy, \text{but} without affecting the primary task. 



The paper is organized as follows. Sec.~\ref{sec:relatedwork} and~\ref{sec:background} reviews and recapitulates prior related works. Sec.~\ref{sec:method} presents details of our approach to manage multiple subtasks subject to redundancy insufficiency. Sec.~\ref{sec:experiment} introduces two case studies
with experimental results to verify the performance of our approach. Sec.~\ref{sec:conclusion} concludes this paper and our future work.









\section{Related Work}\label{sec:relatedwork}

Our work is  in the intersection of \textit{inverse~kinematic control}, \textit{redundancy resolution}  and \textit{prioritized multitasking}.


The very early works of redundant robots have derived the fundamental solution 
to redundancy resolution by using Jacobian pseudoinverse to find the instantaneous relationship between the joint and task velocities.
The later extensive investigations, essentially, have been developed explicitly or implicitly from the Jacobian pseudoinverse via either optimization or task augmentation. Typically, redundancy resolution via optimization incorporates additional additional subtasks or objectives by minimizing certain task-oriented criteria~\cite{ficuciello2015variable,khatib1986real}. For example,  obstacle avoidance is enforced by minimizing a function of artificial potential defined over the obstacle region in configuration space~\cite{ge2002dynamic}.  The task augmentation approaches  address additional subtasks by augmenting an integrated task vector containing all subtasks, where the extended or augmented Jacobians are formulated to enforce additional tasks~\cite{sciavicco1988solution,zanchettin2011general,benzaoui2010redundant}.
 






The majority of frequently applied approaches to redundancy resolution are fundamentally based on the null space projection strategy~\cite{sadeghian2013dynamic,ott2015prioritized,dietrich2018hierarchical}.
In compliance with a dynamically consistent task hierarchy of this line of work, additional subtasks are preformed only in the null space of a certain higher-priority task, typically by successive null space projections~\cite{ott2015prioritized,dietrich2012integration} or augmented null space projections~\cite{slotine1991general, sentis2005synthesis}. We also build our control law upon this technique by performing all subtasks in the null space of the primary task. The aforementioned  
Jacobian pseudoinverse centered approaches, however, work mostly under the premise of sufficient redundancies for multitasking, which instead is the major challenge motivating and addressed by our work.

\text{Our work} is also related to prioritized multitask control, which is mainly focused on addressing task incompatibility by defining suitable evolution of task priorities~\cite{lober2015variance, dehio2019dynamically,penco2020learning}. Typically, priorities are given to safety-critical tasks such as balancing if conflict or incompatibility occurs~\cite{modugno2016learning,di2019handling}. Different from this line of studies, our work mainly focus on the issue of insufficient robot redundancy, and therefore all substasks have to compete for redundancy even in the absence of task incompatibility.

\section{Background}
\label{sec:background}

Our work is built upon prior literature in inverse differential kinematics and null space projection based redundancy resolution.

\subsection{Inverse Differential Kinematics}
Let $\q \in\mathbb{R}^n$ denote the joint configuration of a robot with 
$n$ degrees of freedom. Let $\x \in\mathbb{R}^m$ denote the vector of task variables
in a suitably defined $m$-dimensional task space.
%
The first-order \textit{differential kinematics} is usually expressed as
\begin{equation}\label{eq:differen}
    \dot{\x} =\textbf{\text{J}}(\q) \dot{\q}
\end{equation}
where $\dot{\x}$, $\dot{\q}$ are vectors of task and joint velocities respectively.  $\J(\q)$ is the $m\times n$ Jacobian matrix. 
The dependence on $\q$ is omitted hereafter for notation compactness.

Typically, one has $n \geq m$ for a redundant robot, 
i.e. the robot has a $(n\!-\!m)$-dimensional \textit{redundancy space} for subtasking. 
Then the general \textit{inverse differential kinematics} solution of Eq.~\ref{eq:differen} is usually expressed as 
\begin{equation}\label{eq:inverseKinematics}
     \dot{\q} =\textbf{J}^{+} \dot{\x} +(\textbf{I}- \textbf{J}^{+}\textbf{J} ) \dot{\q}_0 
\end{equation}

\noindent where $\J^{+} \in \mathbb{R}^{n\times m}$ is the pseudoinverse matrix of $\J$. $\textbf{N}(\J)=\textbf{I}-\J^{+}\J \in \mathbb{R}^{n \times n}$ is an operator projecting any arbitrary joint velocity  $\dot{\q}_0 \in \mathbb{R}^n$  into the {null space} of $\J$, i.e. the robot redundancy space.

\subsection{Null Space Projection based Redundancy Resolution}

The projection of  $\dot{\q}_0$  onto the null space ensures no effect on the primary task. Under this premise, the early works \cite{hanafusa1981analysis,maciejewski1985obstacle,nakamura1987task} have proposed the control framework of redundancy resolution with task priority, 
which essentially consists of computing a $\dot{\q}_0$ to suitably enforce a secondary task
in the null space of the primary task.

With reference to Eq.~\ref{eq:inverseKinematics}, the inverse kinematics solution considering a two-order of task priorities (indexed by 1, 2 for the primary and secondary task respectively) can then be expressed as 
\begin{equation}\label{eq:controlLaw_0}
     \dot{\q} = \J^{+}_{1}\dot{\x_{1}} + (\textbf{I}- \J^{+}_1\J_{1} ) [\J_2(\textbf{I}-\J^{+}_1\J_1)]^{+} ( \dot{\x}_2 -\J_2\J^{+}_1 \dot{\x}_1 )
\end{equation}

\noindent where $\dot{\x}_1$, $\dot{\x}_2$ and $\J_1$, $\J_2$ are the task velocities and Jacobian matrices of the primary and secondary task respectively.

\begin{figure}[!t]
	\centering
	\scriptsize
	\def\svgwidth{1\columnwidth}
	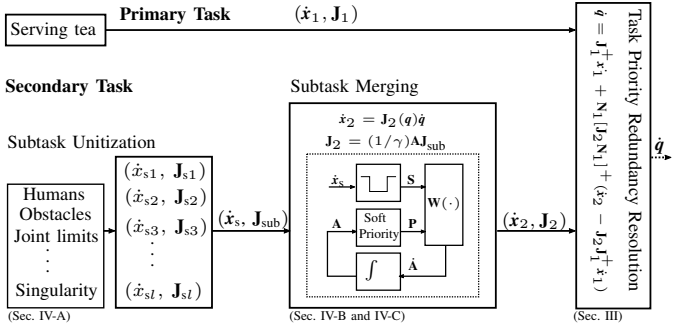
	\caption{Overview of the Approach.}
	\label{fig:overview}
\end{figure}

As illustrated in Fig.~\ref{fig:overview},
we build our control framework upon Eq.~\ref{eq:controlLaw_0}, where we model a virtual dynamic secondary task for subtasks, and then deploy it in the null space of the primary task, such that all subtasks can be suitably executed as good as possible without disturbing 
the primary task.

\section{Method}
\label{sec:method} 

This section presents our approach to manage multiple subtasks subject to redundancy insufficiency (Fig.~\ref{fig:overview}).

\subsection{Subtask Unitization}\label{subsec:a}

We first split and unitize all multi-dimensional subtasks to be executed along with the primary task into a set of one-dimensional \textit{elementary subtasks}. 
For example, the obstacle avoidance of a mobile robot can be unitized into three elementary subtasks of \textit{x}-direction,  \textit{y}-direction and  \textit{yaw}-rotation obstacle avoidance. 
In this manner,  the subtasks can be unitized into a set of elementary subtasks expressed as
\begin{equation}\label{eq:ds}
     \dot{x}_{\text{s}i} = f_i(\bx_{i})\in \mathbb{R} \quad i=1,2,...,l
\end{equation}
\noindent where $l$ is the number of total elementary subtasks.  $\bx_{i}$ is a vector of all related parameters (i.e. the real robot state), and $\dot{x}_{\text{s}i}$ is the desired velocity of the \textit{i}-th elementary subtask. 
Each elementary subtask expressed in the form of Eq.~\ref{eq:ds} need to ensure global stability during construction. Note that the number of elementary subtasks can be less than or equal to the number of redundancies (i.e. $n-m \geq  l$), which implies the robot can provide sufficient redundancies for subtasking.  We focus on the opposite case ($n-m\! < \!l$), where the subtasks have to compete for redundancy due to insufficiency.

The subtask unitization allows our approach to deploy elementary subtasks in a more fine-grained and compact manner, and therefore improve the overall redundancy utilization and availability. Stacking all elementary subtasks together yields a \textit{subtask vector}
\begin{equation}\label{eq:merged}
\begin{split}
        \C=[\dot{x}_{\text{s}1}\: \dot{x}_{\text{s}2}\:...\:\dot{x}_{\text{s}l}]^\text{T}
        =[f_1\;f_2\;...\;f_l]^\text{T}
\end{split}
\end{equation}
Note that we associate an implicit order of elementary subtask priority by index in $\C$, i.e. the smaller the index $i$, the higher the priority of its corresponding elementary subtask.  


Suppose the first-order differential kinematics for the $i$-th elementary subtask is expressed as
\begin{equation}\label{eq:subtaskJo}
    \dot{x}_{\text{s}i} = \J_{\text{s}i}(\q)\dot{\q} 
\end{equation}
\noindent where $\J_{\text{s}i} 
\in \mathbb{R}^{1\times n}$ is its Jacobian matrix.
Substituting Eq.~\ref{eq:subtaskJo} into Eq.~\ref{eq:merged} yields
\begin{equation}\label{eq:elmentary}
    \C = \J_{\text{sub}}(\q)\dot{\q}
\end{equation}
\noindent where $\J_{\text{sub}} = [\J_{\text{s}1}^{\text{T}} \;\J_{\text{s}2}^{\text{T}} \;... \;\J_{\text{s}l}^{\text{T}}]^{\text{T}} \in \mathbb{R}^{l\times n}$ is the merged Jacobian matrix for the elementary subtask set.

\subsection{Merging Subtasks into A Dynamic Secondary Task}\label{subsec:merge}
We then build a \textit{virtual secondary task} $\dot{\x}_2$ from the set of elementary subtasks $\C$ in line with Eq.~\ref{eq:controlLaw_0}
%
\begin{equation}\label{eq:secondary_formal}
    \begin{split}
        \dot{\x}_2 &= \textbf{\text{H}}(\C) \\
    \end{split}
\end{equation}
where ${\textbf{H}(\cdot)}$ is an operator dynamically allocating $n-m$ robot redundancies to $l$ elementary subtasks $\C$ during runtime.

\vspace{2mm}
\noindent\textbf{Multi-Subtask Merging Matrix}: In order to construct the operator ${\textbf{H}(\cdot)}$, we first define  a \text{multi-subtask merging matrix} 
\begin{equation}
 \!  \A(t):=\!
    \begin{bmatrix}
    \alpha_{11}(t) & \alpha_{12}(t) &\!\dotsm &\!\alpha_{1l}(t) \\ \!
    \alpha_{21}(t) & \alpha_{22}(t) &\!\dotsm &\!\alpha_{2l}(t) \\ \!
    \vdots &\vdots  &\!\ddots &\!\vdots\\ \!
    \alpha_{(n-m)1}(t) & \alpha_{(n-m)2}(t) &\!\dotsm  &\!\alpha_{(n-m)l}(t) \!
    \end{bmatrix}\!
\end{equation}
where each entry $\alpha_{ij}$ denotes the weight of the $i$-th redundancy to be allocated to the $j$-th elementary subtask varying w.r.t. time.
It satisfies 
$\sum_{j=1}^l\alpha_{ij}\! = \!\gamma$, where $\gamma \in [0.5, 1]$ is the upper bound for entries in $\A$. The dependence on $t$ is omitted hereafter for notation compactness.
The matrix is initialized with
\begin{equation*}
    \A_0:=\;\left[\gamma \cdot {\bf{I}}_{\left( {n - m} \right) \times \left( {n - m} \right)} \quad  {\bf{0}}_{\left( {n - m} \right)\times \left( {l - n + m} \right)}\right]
\end{equation*}
which implies  the  $\!n\!-\!m$ robot redundancies  will be initially allocated to the first $n-m$ elementary tasks in $\C$, in keeping with the aforementioned implicit indexing task priority.

\vspace{2mm}
\noindent\textbf{Virtual Secondary Task}:
Then the virtual secondary task $\x_2$ is defined as a weighted contributions of  $l$ subtasks as
\begin{equation*}\label{eq:secondary}
     \dot{\x}_2   \! = \!\textbf{\text{H}}(\C)\!=\! (1/{\gamma})\cdot\A_{(n-m)\times l}\dot{\x}_{\text{s}(l \times1)}\!= (1/\gamma)\sum_{j=1}^l{\boldsymbol{\alpha}_{j}\dot{x}_{\text{s}j}  }
\end{equation*}
\begin{equation}\label{eq:secondary2}
    \begin{split}
    \!  & = \! (1/\gamma)\!
        \left[
            \sum_{j=1}^l\alpha_{1j}
            \dot{x}_{\text{s}j}  \:\sum_{j=1}^l\alpha_{2j}
            \dot{x}_{\text{s}j}  \: ...
            \:\sum_{j=1}^l\alpha_{(n-m)j}
            \dot{x}_{\text{s}j}  
        \right]^{\text{T}}   \!\\
       \!& = \! [\dot{x}_{21} \:\:\dot{x}_{21} \:\:\dots  \:\:\dot{x}_{2(n-m)}]^{\text{T}}\!
    \end{split}
\end{equation}
where $\gamma$ acts as a normalizing factor. Eq.~\ref{eq:secondary} also implies  at the \textit{i}-th redundancy, the merging matrix $\A$ dynamically
allocates a virtual task $\dot{x}_{2i}$ characterized by a weighted sum of $l$ elementary subtasks. 

\vspace{2mm}
\noindent
\textbf{Null Space Control}:
Substituting Eq.~\ref{eq:elmentary} and~\ref{eq:secondary} into 
Eq.~\ref{eq:differen} yields
\vspace{-1mm}
\begin{equation}\label{eq:mergedJaco}
        \dot{\x}_2 = \J_2(\q)\dot{\q}
        =(1/\gamma)\A\J_{\text{sub}}(\q)\dot{\q}
\end{equation}
where  $\J_2 = (1/\gamma)\A\J_{\text{sub}}$ is the (merged) Jacobian matrix of the virtual secondary task. 
Then
substituting Eq.~\ref{eq:secondary} and~\ref{eq:mergedJaco} into Eq.~\ref{eq:controlLaw_0} yields our law of redundancy resolution subject to insufficiency 
\begin{equation*}\label{eq:controlLaw_2}
    \qdot \!=\!\J^{+}_1\xdot_1\!+\!
      \textbf{N}_1\J^{\text{T}}_{\text{sub}}\textbf{A}^{\text{T}}(\textbf{A}\J_{\text{sub}}\textbf{N}_1\J^{\text{T}}_{\text{sub}}\textbf{A}^{\text{T}})^{-1}(\A\C \!-\! \A\J_{\text{sub}}\J^{+}_1\xdot_1) \!\!\!\!\\
\end{equation*}
\begin{equation}\label{eq:controlLaw_1}
\textbf{N}_1\!=\textbf{I}-\J_1^{+}\J_1 \in \mathbb{R}^{n \times n} \quad\quad\quad\quad\quad\quad\quad\quad\quad\quad\quad\:\:
\end{equation}
which plays a fundamental role in our control framework. The next section explains how our algorithm dynamically modulates $\dot{\x}_2$ to manage subtasks under this framework.

\subsection{Update of the Merging Matrix}\label{subsec:c}
With reference to Eq.~\ref{eq:secondary}$-$\ref{eq:controlLaw_1}, the dynamic control of multiple subtasks 
relies essentially on the update of $\A$. We formulate an updating strategy to proactively modulate the updating rate of $\A$ by incorporating \textit{task status} and \textit{soft priority} derived heuristically. 

\vspace{2mm}
\noindent
\textbf{Task Status Matrix}:
We define a \text{task status matrix} $\St$ to modulate the updating rate in compliance with task status
\begin{equation}
    \St = \text{diag}(\bar{f}_1,\bar{f}_2,\dots,\bar{f}_l)
\end{equation}
where $\bar{f}_i\in[0,1]$ quantifies the activation status of the $i$-th elementary subtask $\dot{x}_{\text{s}i}$ with a normalized scalar. 
Specifically, if $\dot{x}_{\text{s}i}$ arrives at a stable state,  then there is  $\dot{x}_{\text{s}i}= 0, \bar{f}_i=0$. That is, the \textit{i}-th elementary subtask has been completed and there is no need to assign redundancy to it. On the contrary, if $ \bar{f}_i\rightarrow1$, it indicates the \textit{i}-th elementary subtask is still active and therefore waiting be allocated with a redundancy.

Here we specify $\bar{f}_i$ with the normalizing function  
\begin{equation}\label{eq:normalize}
     \bar{f}_i=1/(1+e^{k_i(d_i+\dot{x}_{\text{s}i})})+1/(1+e^{k_i(d_i-\dot{x}_{\text{s}i})})
\end{equation}
\noindent where ${k_i}$ and ${d_i}$ are the response slope and sensitivity range of the normalizing function. 
%
%
Note one can come up with some other definitions of task status, e.g. one considering the task amplitude. Here we treat all subtasks equally and focus on if an elementary subtask is completed or not.





\vspace{2mm}
\noindent
\textbf{Soft Priority Matrix}:
We derive a \textit{soft priority matrix} $\Pt$ to proactively modulate the updating rate 
\begin{equation}
   \!   \Pt(t):=
    \begin{bmatrix}
        p_{11}(t) & p_{12}(t) &\!\dotsm & \!p_{1l}(t)\! \\
        p_{21}(t) & p_{22}(t) &\!\dotsm &\! p_{2l}(t)\! \\
        \vdots &\vdots  &\!\ddots &\! \vdots\\
        p_{(n-m)1}(t) & p_{(n-m)2}(t) &\!\dotsm  &\! p_{(n-m)l}(t) \!
    \end{bmatrix}
\end{equation}
where each entry $p_{ij}\in (0,1)$ implies a certain value of  \textit{soft priority} proactively modulating the updating rate of the weight $\alpha_{ij}$.
The soft priority is derived 
%
by the following rules\footnote{Note that a \text{dummy row} $\alpha_{0j}=0,\;j=1,2,\dots,l$ and a \text{dummy column} $\alpha_{i0}=0,\;i=1,2,\dots,(n-m)$ are added into $\A$ for the sake of expression simplicity. $\alpha_{0j}=0$ implies the dummy \text{0}-th redundancy will not be assigned to any subtask. $\alpha_{i0}=0$ implies the dummy \text{0}-th task will not occupy any redundancy.}
\begin{equation}\label{eq:rules}
    \begin{split}
        p_{ij} = \prod_{u=0}^{i-1}(1-\alpha_{uj})\prod_{v=0}^{j-1}(1-\alpha_{iv})\prod_{u\neq i}(\gamma-\alpha_{uj})\\
    \end{split}
\end{equation}
for $i=1,2,\dots,(n-m)$ and  $j=1,2,\dots,l$.
%
Each entry $p_{ij}$ extracts implicit soft priority information from $\A$ by explicitly considering the weight distribution over its corresponding redundancy ($i$-th row) and elementary subtask ($j$-th column):
\begin{itemize}
    \item  The term $\prod_{u=0}^{i-1}(1-\alpha_{uj})$ indicates the updating rate of $\alpha_{ij}$ is affected by the weight distribution (for the \textit{j}-th elementary subtask) over the ($i-1$) redundancies previous to the current \textit{i}-th one.  Specifically, given a \textit{j}-th elementary subtask, 
    if its weight at any other redundancy (denoted as \textit{u}-th) previous to the current \textit{i}-th one is close to $\gamma$ (i.e. $\alpha_{uj} \rightarrow \gamma$), it is more likely to be assigned to the \textit{u}-th redundancy. Therefore, the weight at the current \textit{i}-th redundancy will be relatively reduced to proactively quit the competition for the $j$-th  elementary subtask. On the contrary, if its weight at any previous redundancy is close to zero, the weight at the current redundancy will be relatively raised proactively to improve the chance of winning.
    \item The term $\prod_{v=0}^{j-1}(1-\alpha_{iv})$ indicates, symmetrically,
    the updating rate of $\alpha_{ij}$ is affected by the weight distribution (at the $i$-th redundancy) over the $j-1$ elementary subtasks previous to the current \textit{j}-th one. This term decides if the \textit{j}-th elementary subtask should proactively quit or stay in the competition for the \textit{i}-th redundancy.
    \item The term $\prod_{u\neq i}(\gamma-\alpha_{uj})$ acts as a redundancy keeper by  rejecting or zeroing out the weight update at $\alpha_{ij}$ if the $j$-th elementary subtask has been allocated to  any other redundancy (denoted as $u$-th and therefore $\alpha_{uj}=\gamma$) rather than the current $i$-th one. This guarantees the \textit{j}-th elementary subtask will be kept in a redundancy once being allocated to and therefore would not jump back and forth among different redundancies.

\end{itemize}

The soft priority derived above is consistent with the aforementioned indexing priority by explicitly considering the weight distribution over previous redundancies and subtasks. 
It proactively tuning the updating rate 
and therefore leads to a faster convergence speed of the merging matrix $\A$.
Such a prioritizing strategy is aimed at improving the efficiency of redundancy resolution, such that all elementary subtasks can be suitably performed in due course. Note one can come up with some other prioritizing strategies in accordance with context~\cite{penco2020learning,modugno2016learning,di2019handling}.


\vspace{2mm}
\noindent\textbf{Updating the Merging Matrix}:
%
%
%
 %
We define the updating rate $\dot{\A}$ as a combined effect of the task status $\St$ and the soft priority $\Pt$, 
and formulate it based on the winner-take-all strategy\footnote{A detailed explanation of the algorithm and a proof of weight convergence are provided here:  \url{https://github.com/AccotoirDeCola/WinnerTakeAll.} } ${\textbf{W}(\cdot)}$ 
\begin{equation}\label{eq: update_rate}
    \dot{\A} = \textbf{W}(\Pt,\St,\A)
\end{equation}
%
Then the subtask merging matrix $\A$ is updated as follows
\begin{equation}\label{eq: update}
{{\bf{A}}_{{{t + 1}}}} = \max ({\bf{0}},\min ({\gamma\bf{E}},{{\bf{A}}_{{t}}}{{ + }}{{\bf{\dot A}}_{{t}}}\Delta {{t}}))
\end{equation}
where $\bf{E}$ is an all-ones matrix, and $\Delta{t}$ is the update interval.

\section{Experiment Results}
\label{sec:experiment}

This section presents two test cases followed by experimental results to show the performance of our approach.


\begin{figure}[!t]
	\centering
	\tiny
	\def\svgwidth{1\columnwidth}
	\includegraphics[width=3.5in]{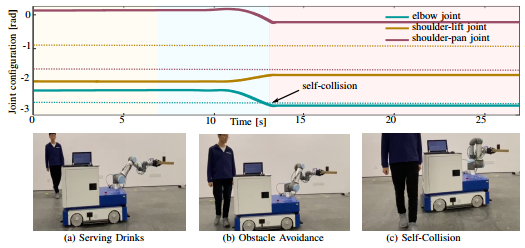}
	\caption{The traditional approach: The robot collides with itself at the elbow joint (the blue line) at around 13s, as the self-collision subtask  is not treated during the whole process due to redundancy insufficiency. Each solid line represents a relevant joint for self-collision avoidance, while the dotted line in the same color represents its joint-collision limit.  }
	\label{fig4:SelfColli}
\end{figure}

\subsection{Experimental Cases}

\noindent\textbf{I. Drink-Serving}: As introduced previously in Fig.~\ref{fig1:intro}, the first test case is about a  mobile robot serving drinks along a desired path. We implement this test case on a real six-DOF UR16e robot manipulator mounted on an omnidirectional mobile platform. Therefore, the robot has in total nine DOFs.  The primary task of serving drinks requires six DOFs and therefore leaves three DOFs as redundancies for subtasking.
The subtasks in this case involve:
\begin{itemize}
    \item A three-dimensional \textit{obstacle-avoidance} subtask, e.g. avoiding the walking human, which can be split into three elementary obstacle-avoidance subtasks.
    \item A three-dimensional \textit{self-collision} avoidance subtask, e.g. avoiding the collision between the manipulator and the platform, which can be split into three elementary self-collision avoidance subtasks.
\end{itemize}

Ideally, both subtasks should be performed simultaneously along with the primary task. However, due to the lack of sufficient redundancies, the six elementary subtasks have to compete for three redundancies during runtime.

\vspace{2mm}
\noindent\textbf{II. Circle-Drawing}: As illustrated in Fig.~\ref{fig4:maintas},  the second case is about a manipulator drawing a circle along a desired end-effector path. We implement this test case using the same robot as the first case, but the mobile platform is fixed at a certain location.  Therefore, the robot has in total six DOFs. The primary task of circle drawing requires three DOFs and therefore leaves three DOFs as redundancies for subtasking. The subtasks in the case involve:

\begin{itemize}
    \item A three-dimensional \textit{singularity-avoidance} subtask, which can be split into three elementary singularity-avoidance subtasks. 
    \item A one-dimensional \textit{wrist-limit} subtask, which simply constraints the wrist joint to a desired angle.
\end{itemize}

Therefore, there are four elementary subtasks competing for three redundancies in this case. 

\subsection{Experimental Results}
We test our approach (Eq.~\ref{eq:controlLaw_1}) on both cases and compare it with the traditional approach (Eq.~\ref{eq:controlLaw_0}). Briefly, given a  case, 
\begin{itemize}
    \item The traditional approach first assigns a number of required DOFs to primary task. Then it allocates the remaining redundancies to as many subtasks as it can  \textit{and} then keep the redundancy allocation. 
    \item The subtask-merging based approach (our approach), as explained in Sec.~\ref{sec:method}, first assigns the required DOFs to  primary task. Then it dynamically allocates the remaining redundancies to all elementary subtasks  generated from subtask unitization  in due time. 
\end{itemize}

\begin{figure}[!t]
	\centering
	\scriptsize
	\def\svgwidth{1\columnwidth}
	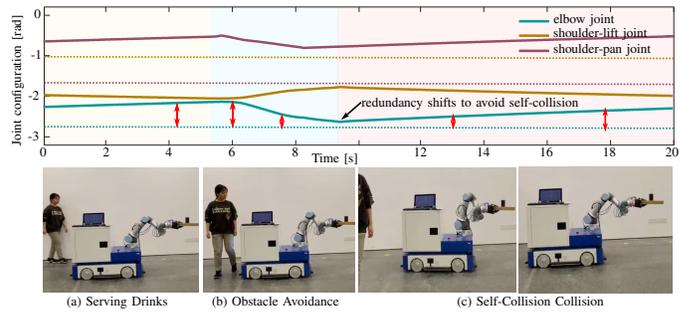
	\caption{Our approach:  
    All subtasks are suitably performed. Three redundancies are first shifted to the obstacle-avoidance subtask to avoid the walking human at round 5s, and then given back to three elementary self-collision subtasks to avoid potential self-collision at around 9s.  }
	\label{fig5:NonSelfColli}
\end{figure}

\begin{figure}[!t]
	\centering
	\tiny
	\def\svgwidth{1\columnwidth}
	\includegraphics[width=3.5in]{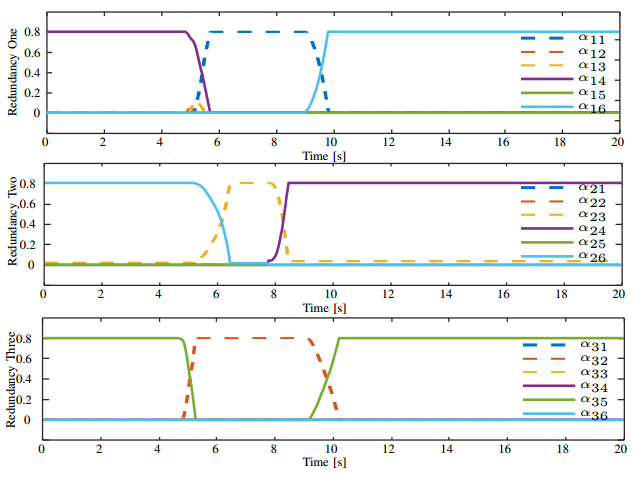}
	\caption{Redundancies shift dynamically among elementary subtasks: 
    Our approach dynamically allocates three redundancies to six elementary subtasks in due course. Each subfigure corresponds to a redundancy, where  three dotted coloured lines correspond to the weights of three elementary obstacle-avoidance subtasks, and  three solid coloured lines correspond to the weights of three elementary self-collision  subtasks, i.e. the update of   $\A$.}
	\label{fig5:WeightMatrix}
\end{figure}

\vspace{2mm}
\noindent\textbf{I. Experimental Results of Drink-Serving:}
Fig.~\ref{fig4:SelfColli} and~\ref{fig5:NonSelfColli} show the results generated respectively by the traditional and our approach during the whole process of the self-collision avoidance subtask. 
Fig.~\ref{fig5:WeightMatrix} shows the redundancy shift among six elementary subtasks (i.e. the evolution of weights in $\A$) generated by our approach. 

\begin{figure*}[!th]
	\begin{center}	
		\mbox{
			\subfigure[Performance on the Singularity-avoidance Subtask ]{{
                    	\centering
                    	\tiny
                    	\def\svgwidth{1\columnwidth}
						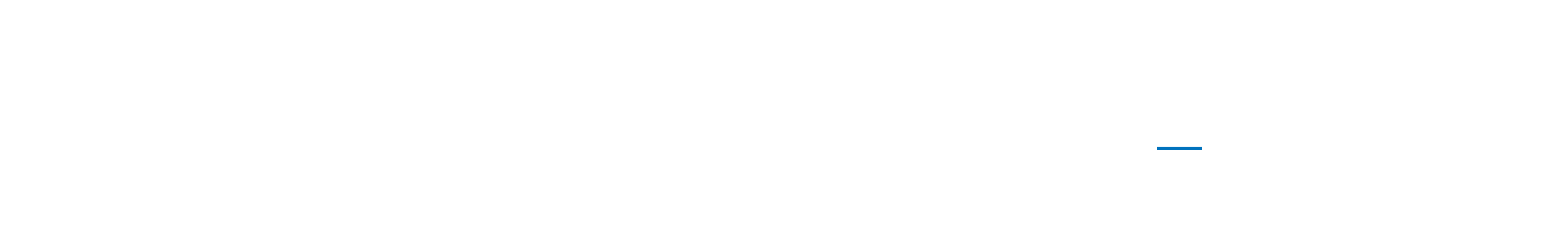
                    	\label{subfig:sin}
					}}
			\subfigure[Performance on the Wrist-Limit Subtask]{{
                    	\centering
                    	\tiny
                    	\def\svgwidth{1\columnwidth}
						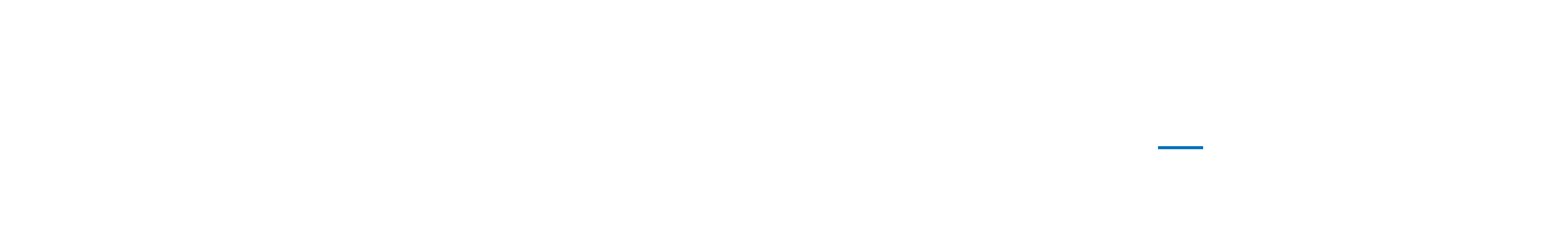
                    	\label{subfig:joint}
					}}					
		}
		\caption{Both approaches perform well in the singularity-avoidance subtask, 
while our approach outperforms in properly addressing the wrist-limit subtask.}
		\label{fig:tc2}
	\end{center}
\end{figure*}

In this case,  as shown in Fig.~\ref{fig4:SelfColli}, the traditional approach allocates three redundancies to the obstacle-avoidance subtask, and then leaves the self-collision subtask untreated since there is no more redundancy available. As a result, 
even though the moving human is successfully avoided during the whole course (as the obstacle-avoidance subtask is taking all redundancies), the robot collides with itself at the elbow joint at around 13 s and locks its manipulator henceforth for mechanical safety, i.e. the robot fails in executing the case.

Instead, as shown in Fig.~\ref{fig5:NonSelfColli} and~\ref{fig5:WeightMatrix}, our approach dynamically allocates three redundancies to six elementary subtasks, and therefore all subtasks are  suitably performed in due course. Specifically, the three redundancies are initially taken by the self-collision subtask, therefore the relative difference between each joint  to its corresponding joint-collision limit (illustrated by the red double-arrowed line segments in Fig.~\ref{fig5:NonSelfColli}) increases in this phase (0s-5s). 
As the human enters the robot's sensing range of obstacle avoidance from around 5s, the  redundancies are shifted to  three self-collision elementary subtasks to keep the robot away from the walking human.
Meanwhile, as a result, the  joint differences for self-collision decrease (but not to zero) till around 9s, when the  redundancies are shifted back to three elementary self-collision subtasks first and last. Accordingly, the joint differences increase again to avoid potential self collisions. All above redundancy shifts can be directly observed in Fig.~\ref{fig5:WeightMatrix}.

Remarkably, Fig.~\ref{fig5:NonSelfColli} and~\ref{fig5:WeightMatrix} also show  redundancy shifts do not need to happen simultaneously, even for the same subtask. That is, our approach allocates redundancies directly to  one-dimensional elementary subtasks rather than their corresponding  high-level   multi-dimensional  subtasks. This is 
thanks to the subtask unitization as introduced in Sec.~\ref{subsec:a}, which greatly improves the redundancy availability and utilization. 
For example, from around 8s to 10s in Fig.~\ref{fig5:WeightMatrix}, the second redundancy is shifted to an elementary self-collision subtask, while the other two redundancies are still occupied by two elementary obstacle-collision subtasks. 
It is also suggested from both figures that the redundancy shift can be performed swiftly (mostly within 1s) and smoothly by our approach.

\vspace{2mm}
\noindent\textbf{II. Experimental Results of Circle-Drawing:}

Fig.~\ref{fig:tc2} shows  results for the second case on the singularity-avoidance and wrist-limit subtasks generated by the traditional and our approach respectively.  Both approaches perform well in the singularity-avoidance subtask (Fig.~\ref{subfig:sin})
while the traditional approach underperforms in the wrist-limit subtask due to redundancy insufficiency (Fig.~\ref{subfig:joint}). 

\begin{figure}[!t]
	\centering
	\tiny
	\def\svgwidth{1\columnwidth}
	\includegraphics[width=3.5in]{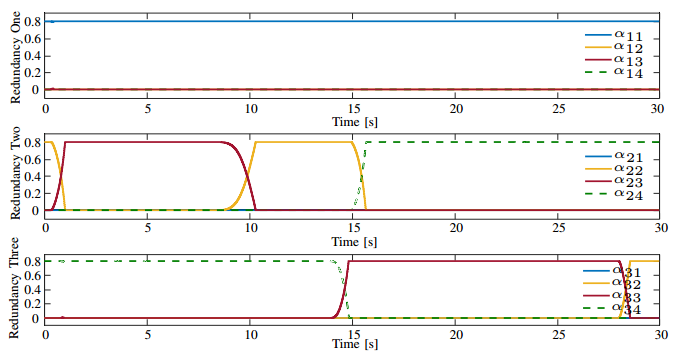}
	\caption{Three redundancies shift dynamically among four elementary subtasks  in due course. The three solid  lines correspond to three elementary singularity-avoidances. The dotted line corresponds to the wrist-limit subtask. }
	\label{fig:2shitf}
\end{figure}

\begin{figure}[!t]
	\centering
	\tiny
	\def\svgwidth{1\columnwidth}
	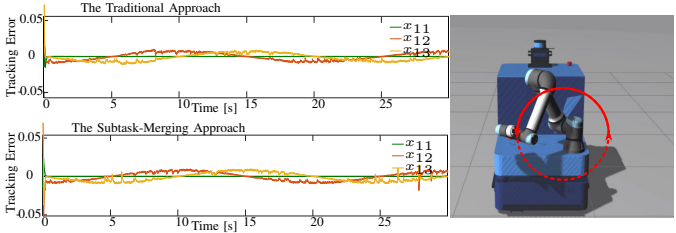
	\caption{The manipulator performs a primary task of drawing a circle along a desired end-effector path.
	Both the traditional and our approach perform well in executing the primary task.   }
	\label{fig4:maintas}
\end{figure}

Fig.~\ref{fig:2shitf} shows the redundancy shifts among four elementary subtasks (i.e. the evolution of $\A$) generated by our approach. Specifically, from 0s to around 9s, two elementary singularity-avoidance subtasks and the wrist-limit subtask are performed. Then at around 9s, the second redundancy is shifted from one elementary singularity-avoidance subtask to the other, i.e. a redundancy shift happens between two elementary subtasks unitized from the same high-level subtask.  
This further proves that our approach allocates redundancies in the elementary subtask level. Such a redundancy shift is in fact due to the change of task status, i.e. a (nearly) completed subtask gives its redundancy to an alive subtask.  

Remarkably, 
Fig.~\ref{fig:2shitf} shows the primary task can be performed well by both approaches, i.e. the primary task is not affected  by the execution of subtasks.
This is thanks to the null space projection technique applied by both approaches.

\section{Conclusion and Future Work}
\label{sec:conclusion}

This work has addressed the constrained redundancy resolution problems where multiple constraints or subtasks in addition to a primary task have to compete  for insufficient redundancies.  The proposed approach based on subtask merging and null space projection
resolves redundancy insufficiency by dynamically allocates them to subtasks in compliance with task status and priority.   
Two real robot case studies with solid and substantial results have proved that our approach  
can be a promising solution 
to suitably handle complex robot applications characterized by dynamic and unstructured environments.
Based on the results, our future works will focus on (1) further modulating and smoothing redundancy shifts to reduce its effect on task execution, e.g. at around 15s in Fig.~\ref{subfig:joint}, the joint difference fluctuates shortly due to a redundancy shift.
and (2) introducing a certain level of predicting capability to the weight updating strategy such as  to proactively predict and accommodate the change of task status, e.g. the occurrence of an emergency.


\section{Appendix}
\label{sec:app}

\subsection{Winner-Take-All Based Updating Algorithm}\label{subsec:WTA}

\begin{algorithm}[!h]
\label{alg:Framwork}  
\caption{$\dot{\A} = \textbf{W}(\A,\St,\Pt)$} 
\hspace*{0.02in} {\bf Input:} 
$\A,\St,\Pt$\\
\hspace*{0.02in} {\bf Output:} 
$\dot{\A}$
\begin{algorithmic}[1]
\State $\dot{\A}=\Pt\times \St $
\For{each row $\dot{\A}_i$ in $\dot{\A}$}
    \State $\omega = \text{argmax}(\dot{\A}_i)$ 
    \If{$\alpha_{i\omega}=\gamma$} 
        \State $\dot{\A}_i\leftarrow \bf{0}$
    \Else
        \State $v = \text{argmax}(\dot{\A}_i- \{ \dot{\alpha}_{i\omega} \} )$
        \State $z = (\dot{\alpha}_{i\omega}+\dot{\alpha}_{iv})/2$
        \State $\dot{\A}_i \leftarrow \dot{\A}_i-z$
        \State $s\leftarrow0$
        \For{$j\leq l$} 
            \If{$\dot{\alpha}_{ij}>0$ and ${\alpha}_{ij}\neq 0$ } 
                \State $s\leftarrow s+\dot{\alpha}_{ij}$
            \EndIf
        \EndFor
        \State $\dot{\alpha}_{i\omega} \leftarrow \dot{\alpha}_{i\omega}-s$
    \EndIf
\EndFor
\State \Return $\dot{\A}$
\end{algorithmic}
\end{algorithm}

Alg.~1 first yields a preliminary updating rate by multiplying the priority matrix $\Pt$ with the task status matrix $\St$ (line 1). Then at each row $\dot{\A}_i$, if the entry $\alpha_{i\omega}$ in $\A_i$ corresponding to the greatest update in $\dot{\A}_i$ is already saturated at 1, then $\A_i$ will not be updated by setting $\dot{\A}_i$ to be $\textbf{0}$ (line 3–5).

Otherwise, the algorithm first lowers $\dot{\A}_i$ to a baseline by subtracting an average of the first-two largest entries (line 7-9). This ensures only one updating rate in $\dot{\A}_i$ is positive, i.e. only one weight $\A_i$ will increase.
Then, in order to ensure the sum of the updating rate is 0, we calculate the sum of the current effective updating rate and  subtract it to the maximum update rate (line 10–14).

\[\begin{array}{l}
s = {{\dot \alpha }_{\omega j}} + T\\
T = \sum\limits_{i \ne \omega  \cap \left( {{{\dot \alpha }_{ij}} > 0 \cup {\alpha _{ij}} \ne 0} \right)} {{{\dot \alpha }_{ij}}} 
\end{array}\]

$S$ of the equation represents the sum of all valid update rates
${\dot \alpha }_{ij}$. 

\[\begin{array}{l}
{{\dot \alpha }_{\omega j}} \leftarrow {{\dot \alpha }_{\omega j}} - S\\
 = {{\dot \alpha }_{\omega j}} - ({{\dot \alpha }_{\omega j}} + T) =  - T
\end{array}\]

\[{{\dot \alpha }_{\omega j}} + \sum\limits_{i \ne \omega  \cap \left( {{{\dot \alpha }_{ij}} > 0 \cup {\alpha _{ij}} \ne 0} \right)} {{{\dot \alpha }_{ij}}}  =  - T + T = 0\]

The above formula indicates that the sum of all valid update rates is 0. Therefore, after the $A$ matrix is updated, the sum of its items remains unchanged.
This will ensure that the weight will not be cleared. 

\subsection{Weight Convergence and System Stability}
This section presents a detailed proof our approach can converge each weight in $A$ to a stable state along both redundancy and subtask.

Suppose two elementary subtasks $f_p$ and $f_q$, where $f_p$ is being (or has been) activated, i.e. $\bar f_p  \simeq  1$, and by contrast $f_q$ is idle, i.e. $\bar f_q  \simeq  0$. 
We aim to prove that the weight transition can be always correctly achieved for both subtasks, such that they can be suitably performed in due course. We open up our proof along the redundancy and subtask space separately.



\vspace{2mm}
\noindent\textbf{I. Weight Transition along Redundancy}: Assume an $i$-th redundancy is available for subtasking $f_p$ and $f_q$.
If in the winner take all process, the winner is $f_p$,
\begin{equation} 
    \Delta \dal_{i-pq} \&:= \dal_{ip}-\dal_{iq}\\
        \&= {\W(\Pt, \St, \A)_{ip}-\W(\Pt, \St, \A)_{iq}}
        \geq 0\\
\end{equation}
The weight will transition from  $f_q$  to $f_p$, and vice versa.\\

If the winner has been born and the maximum update value is still the winner, then the weight of all non-winners is 0, the weight remains stable, and there is no mutual transition (Alg.~1 line 4-5).
If there is a weight transition,the below relationship holds for all $i$ that is not the winner.
\begin{equation} 
    \frac{\W(\Pt, \St, \A)_{ip}-\W(\Pt, \St, \A)_{iq}} {(\Pt\St)_{ip}-(\Pt\St)_{iq}} \geq 1
\end{equation}
In Alg.~1 lines 7 to 9, only the same item z is subtracted from all elements, and the relative distance between elements remains the same. Since neither  $f_q$ nor $f_p$ is winner, there is no action on line 10.

Then the relative updating difference between $f_p$ and $f_q$ is
\begin{equation*}
    \begin{split}
        \Delta \dal_{i-pq} &:= \dal_{ip}-\dal_{iq}  \\
        &= {\W(\Pt, \St, \A)_{ip}-\W(\Pt, \St, \A)_{iq}}
        \geq (\Pt\St)_{ip}-(\Pt\St)_{iq}\\
        &= \prod_{u=0}^{i-1}(1-\alpha_{up})\prod_{v=0}^{p-1}(1-\alpha_{iv})\prod_{u\neq i}(\gamma-\alpha_{up})\fb_p\\
        & - \prod_{u=0}^{i-1}(1-\alpha_{uq})\prod_{v=0}^{q-1}(1-\alpha_{iv})\prod_{u\neq i}(\gamma-\alpha_{uq})\fb_q
    \end{split}
\end{equation*}

\vspace{2mm}
\noindent Specifically, there are four cases:

\vspace{1.5mm}
\noindent\textbf{Case One}: Suppose neither of $f_p$,$f_q$ is occupying a redundancy, i.e. $\alpha_{up}\simeq \alpha_{uq} \simeq 0$, $\forall u\neq i$. Then we have 
\begin{equation*}
    \begin{split}
        0<&\prod_{u=0}^{i-1}(1-\alpha_{up})\prod_{u\neq i}(\gamma-\alpha_{up})\\
        \simeq&\prod_{u=0}^{i-1}(1-\alpha_{uq})\!\prod_{u\neq i}(\gamma-\alpha_{uq}) \simeq \gamma^{n-m-1}
    \end{split}
\end{equation*}
Denote $c=\gamma^{n-m-1}>0$ (a constant), then we have 
\begin{equation*}
    \dal_{i-pq}\! \geq\! (\Pt\St)_{ip}\!-\!(\Pt\St)_{iq}\! \simeq c(\!\prod_{v=0}^{p-1}(1-\alpha_{iv})\fb_p\!-\!\prod_{v=0}^{q-1}(1\!-\!\alpha_{iv})\fb_q)
\end{equation*}

\noindent\textbf{(1)}. If $p<q$, 
we have 
\begin{equation*}
    \dal_{i-pq}\! \geq c\prod_{v=0}^{p-1}(1-\alpha_{iv})(\fb_p-\!\prod_{v=p}^{q-1}(1-\alpha_{iv})\fb_q)
\end{equation*}
which indicates as $\fb_p$ approaches one and $\fb_q$ approaches zero in line with their task status, $\dal_{i-pq}\geq0$ is guaranteed, i.e the weight of $f_p$ will increase relatively faster and therefore a higher task priority is correctly given to $f_p$.

\vspace{1mm}
\noindent\textbf{(2)}. If $p>q$, similarly, we have 
\begin{equation*}
    \dal_{i-pq}\! \geq c\prod_{v=0}^{q-1}(1-\alpha_{iv})(\prod_{v=q}^{p-1}(1-\alpha_{iv})\fb_p-\fb_q)
\end{equation*}
which indicates,
similarly, a higher weight will be eventually transited to $f_p$, as $\fb_p$ and $\fb_q$ vary in accordance with their task status. It also suggests that, however, since $f_q$ is previous to $f_p$ by index, until $\fb_q=0$, $\dal_{i-pq}\geq0$ is not guaranteed. That is, the weight of $f_p$ will not be improved as faster as $f_q$ until $f_q$ is competed, since $f_q$ has a higher indexing priority.



\vspace{1.5mm}
\noindent \textbf{Case Two}: Suppose only $f_p$ is occupying a redundancy, i.e. $\exists u \neq i, \alpha_{up} =\gamma$. 
Then $\dot{\alpha}_{ip} =0$ and therefore $\dal_{i-pq} =  \dal_{ip}-\dal_{iq} = 0 -\dal_{iq}\leq0$. That is, a relatively faster weight increase will be given to $f_q$. This is in compliance with the fact that $\fb_p$ has been allocated with a redundancy and therefore its weight will not increase. A higher weight will be accordingly transited to $f_q$. 

\vspace{1.5mm}
\noindent\textbf{Case Three}: Suppose only $f_q$ is  occupying a redundancy, similarly, we can prove $\dal_{i-pq}  \geq0$ holds, which is consistent with the fact a higher weight is supposed to transit to $f_p$.

\vspace{1.5mm}
\noindent\textbf{Case Four}: Suppose both substasks are  holding redundancies. Then $\dal_{ip} =\dal_{iq} = 0$ and therefore $\dal_{i-pq} = 0$, i.e. there is no relative difference between their updating rate, which is consistent with the fact that subtasks that have been (being) executed will not compete for redundancy and there is no weight transition between them.

\vspace{2mm}
\noindent\textbf{II.$\;$Weight Transition along Subtask}: Suppose the subtask $f_p$ has been allocated to a $u$-th redundancy, i.e. $\exists u, \alpha_{up} =\gamma$. Then at any other $\omega$-th redundancy, it satisfies
$\dot{\alpha}_{\omega p}\leq0, \forall \omega \neq u$.  That is, once a subtask has been allocated at a certain redundancy, the weights of the subtask at other redundancies will not increase, which exactly meets the constraint that an assigned subtask should not jump back and forth.

To sum up, our approach can converge the weights along both the redundancy and the subtask space. Since each subtask controller is stable in design, the entire system can be executed stably once the convergence is achieved.
\bibliographystyle{IEEEtran}
\bibliography{main.bbl}



\vspace{12pt}

\end{document}